\documentclass[11pt,a4paper]{article}
\usepackage{emnlp2020_arxiv}
\usepackage{hyperref}
\usepackage{times}
\usepackage{latexsym}

\usepackage{adjustbox}
\usepackage[normalem]{ulem}

\def\Snospace~{\S{}}
\def\Ssmallspace~{\S{}\,}

\usepackage{pdflscape}

\usepackage{booktabs}
\usepackage{placeins}
\usepackage{amssymb}
\usepackage{amsmath}
\usepackage{bbm}
\usepackage{soul}
\definecolor{darkgreen}{HTML}{006400}
\definecolor{darkgrey}{HTML}{AFAFAF}
\newcommand{\mcL}{\mathcal{L}}
\newcommand{\mcV}{\mathcal{V}}
\newcommand\given{\,|\,}
\usepackage{soul}

\usepackage[normalem]{ulem}

\definecolor{darkgreen}{HTML}{006400}
\definecolor{darkgrey}{HTML}{AFAFAF}

\usepackage{microtype}

\aclfinalcopy % Uncomment this line for the final submission

\newcommand\blfootnote[1]{%
  \begingroup
  \renewcommand\thefootnote{}\footnote{#1}%
  \addtocounter{footnote}{-1}%
  \endgroup
}
\title{SMRT Chatbots: Improving Non-Task-Oriented Dialog\\ with Simulated Multiple Reference Training}

\author{Huda Khayrallah  \\
  Johns Hopkins University \\
  \texttt{huda@jhu.edu} \\\And
  Jo\~ao Sedoc \\
   New York University \\
  \texttt{jsedoc@stern.nyu.edu} \\}

\date{}

\begin{document}
\maketitle
\begin{abstract}
Non-task-oriented dialog models suffer from poor quality and \mbox{non-diverse} responses. To overcome limited conversational data, we apply Simulated Multiple Reference Training \cite[SMRT;][]{khayrallah-etal-2020-simulated}, and use a paraphraser to simulate multiple responses per training prompt. We find SMRT improves over a strong Transformer baseline as measured by human and automatic quality scores and lexical diversity. We also find SMRT is comparable to pretraining in human evaluation quality, and outperforms pretraining on automatic quality and lexical diversity, without requiring related-domain dialog data.\blfootnote{This work will be published at EMNLP 2020.} 

\end{abstract}
\section{Introduction}
Non-task-oriented dialog is a low-resource NLP task.
 While large and noisy related corpora exist \cite[e.g. movie subtitles, social media, and irclogs;][]{serban2018survey},  the publicly-released curated corpora are small.  
\citeauthor{serban2018survey} note that smaller corpora have lower lexical diversity and topic coverage, leading to models with poor quality \mbox{non-diverse} responses.
Pretraining on larger data may improve performance, but requires a large dialog corpus in the right language and related domain. 

We leverage Simulated Multiple Reference Training \cite[SMRT;][]{khayrallah-etal-2020-simulated} to overcome sparse dialog data.
SMRT uses a word-level knowledge distillation-inspired objective and a paraphraser to simulate multiple references per training example.
\citeauthor{khayrallah-etal-2020-simulated}  introduce SMRT for machine translation (MT) and simulate training on \emph{all} translations for a source sentence, assuming: (1)~all paraphrases of a target are translations of the source; and (2)~all translations of the source are paraphrases of the target. (1)~is true for dialog, but (2)~is not---valid chatbot responses vary in meaning. 
SMRT  captures
\emph{syntactic} diversity though it cannot represent all \emph{semantic} variations.

We apply SMRT to chatbots and find that it: (1)~improves human and automatic quality scores; 
(2)~improves lexical diversity; (3)~performs as well as pretraining in human evaluation with better performance on automatic measures of  diversity and quality. 

\begin{table}
\vspace{10pt}
 \begin{adjustbox}{max width=\linewidth}
 \centering
\addtolength{\tabcolsep}{-3pt}
\begin{tabular}{ll}
\toprule
prompt:&Study, study, study. I want to learn a lot.\\
response:&You are going to take courses?            
 \\
\midrule
 paraphrases:&\\
\multicolumn{2}{r}{\includegraphics[width=3in]{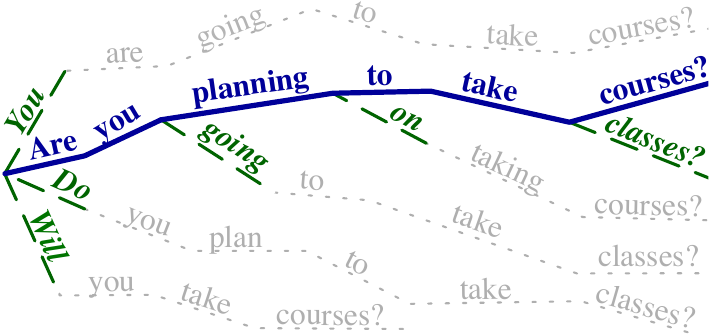}}\\
\bottomrule
\end{tabular}
 \end{adjustbox}
\caption{A DailyDialog training pair and  paraphrases.
 The tree of paraphrases includes some \textcolor{darkgrey}{\dotuline{possible paraphrases}} of the original promt, a \textcolor{darkblue}{\uline{\textbf{sampled path}}} and some of the other   \textcolor{darkgreen}{\textbf{\textit{\dashuline{tokens also considered in the training objective}}}}.}
\label{tab:para_example}
\end{table}

\section{Method}
We model the non-task-oriented dialog system (chatbot) task as conditional language modeling. These models are typically trained using Negative Log Likelihood (NLL) with respect to a single reference.
An alternative approach is Knowledge Distillation \cite[][]{hinton2015distilling,kim-rush-2016-sequence} which assumes access to a teacher distribution ($q(y \given x)$) and minimizes the cross entropy with the teacher's probability distribution.

\paragraph{Simulated Multiple Reference Training}
SMRT is structured similarly to word-level Knowledge Distialltion, but uses a paraphraser as the teacher distribution ($q(y' \given y)$). The paraphraser conditions on the reference $y$ (rather than the source $x$) and generates a paraphrase $y'$. 
Additionally, SMRT \emph{samples} a new paraphrase of the reference every epoch. 
The SMRT training objective for the $i^{th}$ target word in the reference $y$, given the prompt $x$, with a target vocabulary $\mcV$ is:
\begin{align*}\label{eq:para}
\mcL_{\text{SMRT}}=
-\;\sum_{v \in \mcV}^{} \Big[ \;  p_{\textsc{paraphraser}}&(y'_i=v \given y,y'_{j<i}) \\ \nonumber
\times \; \log  \big( p_{\textsc{chatbot}}&(y'_i=v \given x,y'_{j<i})\big)\Big]  \nonumber
\end{align*}
The paraphraser and chatbot each condition on the previously sampled paraphrase tokens ($y'_{j<i}$).

\section{Experimental Setup}
\label{sec:experiment}
\subsection{Dialog models}
\label{sec:dialog_models}
We train Transformer \cite{transformer} chatbots in \textsc{fairseq} using parameters from the \textsc{flores}\footnote{\href{https://github.com/facebookresearch/flores}{\texttt{github.com/facebookresearch/flores}}} benchmark for low-resource MT \cite{guzman-etal-2019-flores} for both a standard NLL baseline and SMRT.\footnote{\href{https://github.com/thompsonb/fairseq-smrt}{\texttt{github.com/thompsonb/fairseq-smrt}}}
Following \citet{khayrallah-etal-2020-simulated}, we sample from the 100 highest probability tokens from the paraphraser distribution at each time-step \cite{fan-etal-2018-hierarchical}.

We train and evaluate on DailyDialog \cite{li-etal-2017-dailydialog}, a high quality corpus with multiple references for evaluation. We train on the $\sim\,$80,000 turns of English-learners practicing  `daily dialogues' in various contexts, e.g., chatting about vacation or food.

See \autoref{app:expt} for full details for replication.

\subsection{Paraphraser}
\label{sec:pp}
We use the state-of-the-art \textsc{Prism} multilingual paraphraser \citet{thompson-post-2020-automatic,thompson-post-2020-paraphrase}.\footnote{\href{https://github.com/thompsonb/prism}{\texttt{github.com/thompsonb/prism}}} 
It is trained as a multilingual MT model on $\sim\,$100 million sentence pairs in 39 languages. Paraphrasing is treated as  zero-shot translation (e.g., English to English).

\subsection{Evaluation Protocols}
\textbf{Human Evaluation}
We use Amazon Mechanical Turk
to collect human judgments. For every HIT we display a prompt and two responses; 
the worker indicates their preferred response (or tie). 
Following \citet{baheti-etal-2018-generating}, we employ the pairwise bootstrap test \citep{efron1994introduction} and report statistical significance at the 95\% confidence level.

\paragraph{Automatic Quality Evaluation}
\label{sec:eval_auto_q}
We use  \textsc{multirefeval} for DailyDialog \cite{gupta-etal-2019-investigating}. 
In \autoref{sec:results} we report \textsc{METEOR}, \textsc{ROUGE-L}, and \textsc{Greedy Match} for the original and multiple references. See \autoref{app:results} for all 14 metrics.
For reading ease we report metrics scaled 0 to 100.

 \paragraph{Automatic Diversity Evaluation}
\label{sec:eval_auto_d}
To measure lexical diversity, we use the type/token ratio of unigrams, bigrams, and trigrams \cite{li-etal-2016-diversity}. 

\section{Results}
\label{sec:results}

\label{sec:results_human}
SMRT is preferred over the baseline system in human evaluation, as shown in \autoref{tab:main_human}. 
\label{sec:results_automatic}
It outperforms the baseline in automatic quality too: see  
\autoref{tab:main_automatic_results}. 
Our \emph{baseline} outperforms nearly all systems in \citet{gupta-etal-2019-investigating} for these metrics,\footnote{Except CVAE on single reference METEOR.} 
suggesting it is a strong baseline.
SMRT has higher lexical diversity than the baseline, though not as high as the human reference response (\autoref{tab:main_lexical_diversity}).

\begin{table}[ht!]
\centering
\begin{tabular}{ccc}
\toprule
baseline & SMRT & tie \\
\midrule
35.8\% & 	{\bf43.5}\% &	20.6\%\\
\bottomrule
\end{tabular}

\caption{Human preference judgments. The output of SMRT is preferred over the baseline system. This preference is statistically significant at the 95\% confidence level.}
\label{tab:main_human}
\end{table}

\begin{table}[ht!]
\centering
\addtolength{\tabcolsep}{-2pt}
\begin{tabular}{l|ccc|ccc}
\toprule
 & \multicolumn{3}{c}{Multi-Ref} & \multicolumn{3}{|c}{Single-Ref}\\\midrule
         & M & R & GM  & M & R & GM \\ \midrule
baseline & 12.8  & 34.0  &   76.9               & 6.9  &20.9    & 71.2              \\
SMRT   &\bf 13.8  & \bf 36.1    &\bf 77.7               & \bf8.1  & \bf24.0    & \bf72.5 \\
\bottomrule
\end{tabular}

\caption{SMRT outperforms the baseline on \textsc{METEOR} (M), \textsc{ROUGE} (R), and \textsc{Greedy Match} (GM) for single and multi-reference scoring. }

\label{tab:main_automatic_results}
\end{table}

\begin{table}[ht!]
\centering
\addtolength{\tabcolsep}{-2pt}
\begin{tabular}{l|ccc}
\toprule
                                   & 1-grams & 2-grams & 3-grams \\\midrule
human reference   & 6.3\%     & 38.9\%    & 72.7\%    \\
\midrule
baseline  & 2.9\%     & 11.6\%    & 20.4\%    \\
SMRT & {\bf3.8}\%     &  {\bf17.4}\%    &  {\bf32.2}\%   \\
\bottomrule

\end{tabular}

\caption{Type/Token ratio for the baseline and SMRT. SMRT has higher lexical diversity than the baseline.}

\label{tab:main_lexical_diversity}
\end{table}
 \pagebreak

\section{Analysis}
SMRT outperforms a strong baseline; here we analyze it in 
additional settings: pretraining and MMI. 

\subsection{Pretraining}
Pretraining is another way of incorporating auxiliary data in the model. 
We pretrain on the OpenSubtitles corpus \cite[OS;][]{lison-tiedemann-2016-opensubtitles2016},\footnote{\href{https://opensubtitles.org}{\texttt{opensubtitles.org}}}
which consists of $\sim\,$200 million turns from movie subtitles. Similar to DailyDialog, it consists of conversational data on a variety of topics. 
After pretraining on OS, we  fine-tune on DailyDialog. 
\paragraph{Results}

In the human evaluation (\autoref{tab:pretrain_human}), SMRT performs comparably to baseline pretraining. In automatic evaluation (\autoref{tab:pretrain_automatic_results}), SMRT outperforms pretraining.
We combine SMRT with pretraining\footnote{We pretrain with NLL then fine-tune with SMRT.} and find that this again performs comparably to baseline pretraining in human evaluation, and pretraining with SMRT performs better in the automatic evaluation.
Finally, we compare SMRT with and without pretraining, and find with pretraining is preferred in human evaluation, while they perform similarly on the automatic metrics. 

Pretraining improves the NLL baseline's diversity, but SMRT's diversity is still better. 
Combining SMRT with pretraining improves diversity compared to pretraining alone: see \autoref{tab:pretrain_lexical_diversity}.

Overall, SMRT performs on par with pretraining in terms of human evaluation of quality, with better diversity and automatic metrics of quality.\footnote{We hypothesize a conversation-level evaluation would further highlight the strengths of SMRT, by allowing for human judgments of diversity  but that is beyond our budget.}

\paragraph{Discussion}
It can be hard to find  \emph{dialog} corpora that are \emph{large}, \emph{domain relevant}, and \emph{in-language}.

Unlike pretraining, SMRT incorporates non-dialog data.  
\textsc{Prism} 
 was trained to translate, and leveraged as a paraphrase model using zero-shot translation. It is not trained to generate dialog, yet we still leverage it to improve a chatbot.

\begin{table}[t!]
\centering
\begin{adjustbox}{max width=\linewidth}
\addtolength{\tabcolsep}{-3.pt}
\begin{tabular}{ll|ccc}
\toprule
M1                  & M2             & M1 & M2         & tie    \\ \midrule
PT + baseline  & SMRT                & 31.6\%  & {32.7}\% & 35.8\% \\
PT  + baseline & PT  + SMRT & 34.9\%  & {36.3}\% & 28.8\% \\
SMRT& PT + SMRT & 32.3\%  & {37.4}\% & 30.2\% \\ \bottomrule
\end{tabular}
\end{adjustbox}
\vspace{1mm}
\caption{Pretraining (PT) human preferences. SMRT and NLL pretraining perform comparably, adding SMRT to pretraining is comparable to NLL pretraining, and pretrained SMRT outperforms SMRT alone. None of the preferences between models in this table are statistically significant at the 95\% confidence level.}
\vspace{3mm}

\label{tab:pretrain_human}
\end{table}
\begin{table}[ht!]
\centering
\addtolength{\tabcolsep}{-2.5pt}
\begin{tabular}{l|ccc|ccc}
\toprule
 & \multicolumn{3}{c}{Multi-Ref} & \multicolumn{3}{|c}{Single-Ref}\\\midrule
         & M & R & GM  & M & R & GM \\ \midrule
baseline & 12.8  & 34.0  &   76.9               & 6.9  &20.9    & 71.2               \\
SMRT   & 13.8  & 36.1    & \bf77.7               & \bf8.1  & \bf24.0    & \bf72.5 \\
\midrule
PT + baseline &13.6 & 35.8 & 77.5 & 7.1 & 21.7 & 71.5 \\
PT + SMRT   & \bf13.9 & \bf36.6 & 77.6 & 7.9 & 23.7 & 72.3
\\ \bottomrule
\end{tabular}
\vspace{1mm}

\caption{SMRT alone outperforms baseline pretraining (PT)  on  \textsc{METEOR} (M), \textsc{ROUGE} (R), and \textsc{Greedy Match} (GM) for single and multi-reference scoring.}
\vspace{3mm}

\label{tab:pretrain_automatic_results}
\end{table}
\begin{table}[ht!]
\centering
\addtolength{\tabcolsep}{-2pt}
\begin{tabular}{l|ccc}
\toprule
 & 1-grams & 2-grams & 3-grams \\\midrule
human reference   & 6.3\%     & 38.9\%    & 72.7\%    \\
\midrule
baseline  & 2.9\%     & 11.6\%    & 20.4\%    \\
SMRT & 3.8\%     &  17.4\%    & {\bf32.2}\%   \\ \midrule
PT + baseline &3.5\%     & 14.2\%    & 24.6\%    \\
PT + SMRT & {\bf4.1}\%     & {\bf17.9}\%    &31.8\%     \\\bottomrule
\end{tabular}
\vspace{1mm}

\caption{Type/Token ratios for pretraining (PT). Pretraining the baseline model increases lexical diversity. SMRT and PT SMRT outperform the baseline and PT + baseline in lexical diversity by similar amounts. }
\vspace{3mm}

\label{tab:pretrain_lexical_diversity}
\end{table}

 The paraphraser is trained on less data ($\sim\,$100 million sentences pairs, with $\sim\,$17 million English sentences) than is used for OpenSubtitles pretraining ($\sim\,$200 million turns---all in English), thus competitive performance is not a result of more data.

\textsc{Prism} was trained on formal text: Wikipedia, news (Global Voices, and SETimes) parliamentary proceedings (EuroParl), and documents (United Nations), not casual or conversational data. 
So SMRT  works in spite of a domain mismatch with the paraphraser, which is especially useful when there is no domain-relevant dialog pretraining data.

While dialog research currently focuses on English, \textsc{Prism} has coverage of 39 languages meaning SMRT can be applied in other languages. 

DailyDialog is well matched to OpenSubtitles, and yet SMRT performs as well as pretraining on OS. This suggests SMRT is effective at leveraging non-dialog data, which is crucial when no in-domain, in-language dialog data is available.

\begin{table*}
 \begin{adjustbox}{max width=\linewidth}
 \centering
\addtolength{\tabcolsep}{-3pt}
\begin{tabular}{lll}
\toprule
prompt:&(a)~~Listen, Karen, I need your help.&(b)~~I try my best to eat only fruits,\\
&\phantom{(a)~~}I don't know anyone here yet.&\phantom{(b)~~}vegetables, and chicken.\\\midrule
baseline:&\phantom{(a)~~}What's wrong with me?&\phantom{(b)~~}I don't know what to eat.\\
SMRT:&\phantom{(a)~~}I am glad to help you, what is wrong?&\phantom{(b)~~}What kind of fruit do you prefer?\\
PT + baseline:&\phantom{(a)~~}I'm sorry to hear that.&\phantom{(b)~~}What kind of vegetables do you like?\\
PT + SMRT:&\phantom{(a)~~}I am sorry, Karen, I can't help you.&\phantom{(b)~~}What kind of food do you eat?\\
\bottomrule
\end{tabular}
\end{adjustbox}
\caption{Two example evaluation prompts, with various system outputs. SMRT outputs are better than the baseline. }
\label{tab:dialog_example}
\end{table*}

\subsection{MMI}
\begin{table}[t]
\centering
\begin{tabular}{ccc}
\toprule
baseline + MMI & SMRT + MMI & tie \\
\midrule
34.7\% & {\bf38.4}\%&	26.9\%\\
\bottomrule
\end{tabular}
\caption{Human preferences judgments. When comparing models with MMI decoding, SMRT is preferred. This preference is statistically significant at the 95\% confidence level.}
\label{tab:mmi_human}
\end{table}
\begin{table}[t]
\centering
\addtolength{\tabcolsep}{-2.5pt}
\begin{adjustbox}{max width=\linewidth}
\begin{tabular}{l|ccc|ccc}
\toprule
 & \multicolumn{3}{c}{Multi-Ref} & \multicolumn{3}{|c}{Single-Ref}\\\midrule
         & M & R & GM  & M & R & GM \\ \midrule
baseline & 12.8  & 34.0  &   76.9               & 6.9  &20.9    & 71.2               \\
SMRT   & \bf13.8  &\bf 36.1    & \bf77.7               & \bf8.1  & \bf24.0    & \bf72.5 \\         
\midrule
baseline + MMI & 12.7 & 33.5 & 76.7 & 6.6 & 20.1 & 70.8 \\
SMRT + MMI   & 13.7 & 35.8 & 77.6 & 7.9 & 23.5 & 72.3
\\ \bottomrule
\end{tabular}
\end{adjustbox}
\caption{MMI degrades both baseline and SMRT performance on \textsc{METEOR} (M), \textsc{ROUGE} (R), and \textsc{Greedy Match} (GM) for single and multi-ref scoring. SMRT + MMI  still outperforms baseline + MMI. }\label{tab:mmi_automatic_results}
\end{table}
\begin{table}[t]
\centering
\addtolength{\tabcolsep}{-2pt}
\begin{tabular}{l|ccc}
\toprule
 & 1-grams & 2-grams & 3-grams \\\midrule
human reference   & 6.3\%     & 38.9\%    & 72.7\%    \\
\midrule
baseline  & 2.9\%     & 11.6\%    & 20.4\%    \\
SMRT & {\bf3.8}\%     &  {\bf17.4}\%    &  {\bf32.2}\%   \\ \midrule

baseline + MMI  & 2.9\%     & 10.1\%    & 17.5\%    \\
SMRT + MMI  & 3.6\%     & 15.7\%    & 28.6\%     \\\bottomrule

\end{tabular}
\caption{Type/Token ratio comparison with MMI. MMI degrades lexical diversity for both methods.}
\label{tab:mmi_lexical_diversity}
\end{table}

Maximum Mutual Information (MMI) decoding,
$(1 - \lambda) \log p(y|x) + \lambda \log p(x|y)$, is commonly used in dialog to increase response diversity \cite{li-etal-2016-diversity}, however we did not find it helpful in our experiments.  
Following MMI-bidi, we rerank a 100-best list with a reverse model.%
\footnote{We sweep $\lambda$ of 0.1, 0.2, 0.3, 0.4, 0.5. $0.1$ performs best on the automatic quality metrics, so we use that for analysis.}
When comparing both models with MMI, we find humans prefer SMRT to the baseline, see \autoref{tab:mmi_human}. 
MMI degrades automatic measures of quality (\autoref{tab:mmi_automatic_results}) and diversity (\autoref{tab:mmi_lexical_diversity}) of both the baseline and SMRT models compared to standard decoding. The quality degradation is similar for both, but the degradation in diversity is more pronounced for SMRT.

\subsection{Examples}
For a training pair and paraphrased responses, see \autoref{tab:para_example}.
SMRT decreases that number of dull and off-topic answers, see \autoref{tab:dialog_example}.
In prompt (a), the baseline is off-topic. Pretraining expresses sympathy, but is unhelpful. SMRT and pretrained SMRT give relevant responses. 
In (b), the baseline has the right general topic but is a poor response. Both SMRT variants and the pretrained baseline respond well.
For more examples, see \autoref{app:ex}. 

\section{Related work}
\paragraph{Paraphrasing}
Neural paraphrasing is actively improving \cite{wieting-etal-2017-learning,wieting-etal-2019-simple,li-etal-2018-paraphrase,wieting-gimpel-2018-paranmt,hu-etal-2019-improved,parabank,hu-etal-2019-large,thompson-post-2020-automatic,thompson-post-2020-paraphrase};
we expect future improved paraphrasers will improve SMRT.

\paragraph{Simulated Multiple Reference Training}
\citeauthor{khayrallah-etal-2020-simulated} use a paraphraser trained on \textsc{Parabank2} \cite[an English paraphrase dataset created using back-translation;][]{hu-etal-2019-large} for SMRT. \citet{thompson-post-2020-automatic} introduced \textsc{Prism}; they show  \textsc{Prism} outperforms \textsc{Parabank2} and that \textsc{Parabank2} is biased against producing the input as the paraphrase, while \textsc{Prism} is not. 
Thus, while \citeauthor{khayrallah-etal-2020-simulated} use a SMRT objective 
with 50\% probability and standard NLL otherwise, we only use SMRT.
\pagebreak
\paragraph{Paraphrastic Dialog Augmentation}
There is little work on data augmentation for chatbots, but there is a variety of work on \emph{task-oriented dialog} augmentation. 
\citet{Kurata+2016} use self-training with noisy decoding to create additional target side data. 
Using a seq-to-seq model, \citet{hou-etal-2018-sequence}  generate diverse lexical and syntactic alternatives within a semantic frame.
 \citet{gao2020paraphrase} jointly train a paraphrase model and a response generation model
using dialog data. 
These works generate paraphrases using dialog training data; in contrast, we leverage additional corpora. 
\citet{niu-bansal-2018-adversarial,niu-bansal-2019-automatically} include paraphrasing as one of several augmentation policies, using external paraphrase data. 
For other NLP tasks, \citet{hu-etal-2019-improved} perform paraphrastic data augmentation for natural language inference, question answering and machine translation.

\paragraph{Diversity}
A variety of decoding approaches address diversity in chatbot output, including: MMI  \cite{li-etal-2016-diversity},
various random sampling (e.g. \citet{fan-etal-2018-hierarchical}), modified beam search \cite{cho2016noisy,vijayakumar2016diverse,tam2019cluster,kulikov-etal-2019-importance}
 and over-generating and clustering post-decoding \cite{ippolito-etal-2019-comparison}.
In this work we improve training, which can be combined with any decoding strategy.

\citet{NIPS2018_7452} and \citet{ xu-etal-2018-diversity} use adversarial training to encourage diversity. 
\citet{ippolito-etal-2019-comparison} note such methods are `task-specific and difficult to implement.'
SMRT is general with simple public code.
\citet{jiang-de-rijke-2018-sequence} connect the low-diversity problem to overconfidence in the model distribution. Since it trains toward a distribution rather than a 1-hot vector, SMRT may have more reasonable confidence levels.
\section{Conclusion}
SMRT improves upon a strong Transformer baseline in quality and diversity. 
It also has human evaluation quality comparable to pretraining, with better automatic quality and lexical diversity.
This method, which works even in settings
where pretraining is impractical due to a lack of \emph{in-domain} \emph{same language} \emph{dialog} data, has a high potential for impact in creating chatbots for more languages.
\pagebreak
\section*{Acknowledgments}
We  thank Patrick Xia, Claire Daniele, Nathaniel Weir,  Carlos Aguirre, and additional anonymous proofreaders 
 for their helpful comments and feedback on the paper. We additionally thank the reviewers for their insightful comments.
 
This work was partially supported by the Amazon AWS Cloud Credits for Research program. This work was supported in part by DARPA KAIROS
(FA8750-19-2-0034). The views and conclusions contained in this work are those of the authors and should not be interpreted as representing official policies or endorsements by DARPA or the U.S. Government.

\bibliographystyle{acl_natbib}
\bibliography{anthology,emnlp2020}
\clearpage
\appendix

\section{Experiment Setup}
\label{app:expt}
\label{app:dialog_models}
\subsection{Dialog Models}
We train Transformer conditional language models in \textsc{fairseq} using parameters from the \textsc{flores}\footnote{\url{https://github.com/facebookresearch/flores/tree/5696dd4ef07e29977d5690d2539513a4ef2fe7f0}} benchmark for low-resource machine translation \cite{guzman-etal-2019-flores} for both the baseline and SMRT.
We use the publicly released SMRT fork of \textsc{fairseq} \cite{ott-etal-2019-fairseq,khayrallah-etal-2020-simulated},\footnote{\url{https://github.com/thompsonb/fairseq-smrt/tree/fdaad1faa01a630beba0c969bd26b65941787752}} along with the \textsc{Prism m39v1} paraphraser \cite{thompson-post-2020-automatic}.\footnote{\url{https://github.com/thompsonb/prism/tree/d2c94b1160f76b3a817eba7f9aba3436deb44731}}

We use a  $5$-layer encoder and decoder, $512$ dimensional embeddings, and $2$ encoder and decoder attention heads. We regularize with $0.2$ label smoothing, and $0.4$ dropout. 
We optimize using Adam with a learning rate of $10^{-3}$. 
We train 100 epochs, and select the best checkpoint based on validation set perplexity. We generate with a beam size of $10$, and no length penalty. 

\autoref{fig:smrt_train} shows the train command for SMRT, \autoref{fig:nll_train} shows the train command for the NLL baseline.

\begin{figure*}[h]
\begin{verbatim}
python fairseq-smrt/train.py \
 $DATADIR \
 --source-lang src \
 --target-lang tgt \
 --seed 10 \
 --save-dir $SAVEDIR --paraphraser-lang-prefix "<en>" \
 --patience 50 --criterion smrt_cross_entropy \
 --paraphraser-model prism/m39v1/checkpoint.pt  \
 --paraphraser-data-dir prism/m39v1/ \
 --paraphraser-sample-topN 100 \
 --prob-use-smrt 1.0 \
 --label-smoothing 0.2 \
 --share-all-embeddings \
 --arch transformer  --encoder-layers 5 --decoder-layers 5 \
 --encoder-embed-dim 512 --decoder-embed-dim 512 \
 --encoder-ffn-embed-dim 2048 --decoder-ffn-embed-dim 2048 \
 --encoder-attention-heads 2 --decoder-attention-heads 2 \
 --encoder-normalize-before --decoder-normalize-before \
 --dropout 0.4 --attention-dropout 0.2 --relu-dropout 0.2 \
 --weight-decay 0.0001 \
 --optimizer adam --adam-betas '(0.9, 0.98)' --clip-norm 0 \
 --lr-scheduler inverse_sqrt --warmup-updates 4000 --warmup-init-lr 1e-7 \
 --lr 1e-3 --min-lr 1e-9 --no-epoch-checkpoints \
 --max-tokens 4000 \
 --max-epoch 100 --save-interval 10 --update-freq 4 \
 --log-format json --log-interval 100 
\end{verbatim}
\caption{SMRT training command.}
\label{fig:smrt_train}
\end{figure*}

\begin{figure*}[h]
\begin{verbatim}
python fairseq-smrt/train.py \
 $DATADIR \
 --source-lang src \
 --target-lang tgt \
 --seed 10 \
 --save-dir $SAVEDIR \
 --patience 50 --criterion label_smoothed_cross_entropy \
 --label-smoothing 0.2 \
 --share-all-embeddings \
 --arch transformer  --encoder-layers 5 --decoder-layers 5 \
 --encoder-embed-dim 512 --decoder-embed-dim 512 \
 --encoder-ffn-embed-dim 2048 --decoder-ffn-embed-dim 2048 \
 --encoder-attention-heads 2 --decoder-attention-heads 2 \
 --encoder-normalize-before --decoder-normalize-before \
 --dropout 0.4 --attention-dropout 0.2 --relu-dropout 0.2 \
 --weight-decay 0.0001 \
 --optimizer adam --adam-betas '(0.9, 0.98)' --clip-norm 0 \
 --lr-scheduler inverse_sqrt --warmup-updates 4000 --warmup-init-lr 1e-7 \
 --lr 1e-3 --min-lr 1e-9 --no-epoch-checkpoints \
 --max-tokens 4000 \
 --max-epoch 100 --save-interval 10 --update-freq 4 \
 --log-format json --log-interval 100 
\end{verbatim}
\caption{Baseline NLL training command.}
\label{fig:nll_train}
\end{figure*}

We train and evaluate on the DailyDialog corpus \cite{li-etal-2017-dailydialog}, as released by ParlAI \cite{alex2017parlai}.\footnote{\url{https://github.com/facebookresearch/ParlAI/tree/1e905fec8ef4876a07305f19c3bbae633e8b33af}} 
We pretrain on the OpenSubtitles corpus \cite[OS;][]{lison-tiedemann-2016-opensubtitles2016}.\footnote{\url{https://www.opensubtitles.org}}

Since SMRT compares the distribution over tokens from the paraphraser and chatbot their vocabularies must match, so we apply the \textsc{Prism} SentencePiece model \cite{kudo-richardson-2018-sentencepiece} 
to the DailyDialog and OpenSubtitles corpora. The ParlAI release of DailyDialog is tokenized and lowercased. Since the data the paraphraser is trained on is not, we detokenize and recase the DailyDialog data. We then provide the \textsc{Prism} dictionary when running \textsc{fairseq-preprocess} (see \autoref{fig:preprocess}).

\begin{figure*}[h]
\begin{verbatim}
python fairseq-smrt/preprocess.py \
--source-lang src --target-lang tgt \
--trainpref $path_to_sentencepieced_data/train.sp \
--validpref $path_to_sentencepieced_data/valid.sp \
--testpref $path_to_sentencepieced_data/test.sp \
--srcdict prism/m39v1/dict.tgt.txt \
--tgtdict prism/m39v1/dict.tgt.txt \
--destdir $databin
\end{verbatim}
\caption{fairseq-preprocess command.}
\label{fig:preprocess}
\end{figure*}

For MMI we use SMRT for the reverse model as well.
For pretraining + SMRT  we use standard NLL for pretraining on OpenSubtitles, and fine-tune on DailyDialog with SMRT. 
\subsection{Evaluation Protocols}
\subsubsection{Human Evaluation}
\label{app:human_eval}
We randomly sample 500 prompt-response pairs from the test set, and filter out any that are not distinct, leaving 482  
pairs.

\subsubsection{Automatic Quality Evaluation}
\label{app:auto_eval}
In \autoref{app:results} we report the full automatic evaluation results of the 14 metrics across both the single reference and multi-reference evaluation from the the multi-reference automatic evaluation framework for DailyDialog released by \citet{gupta-etal-2019-investigating}, which is computed using \textsc{nlg-eval}\footnote{\url{https://github.com/Maluuba/nlg-eval/tree/846166566bf0fdccbaa9e5b41da97147470b525b}} \cite{sharma2017nlgeval}. 
This include word-overlap metrics: BLEU \cite{papineni-etal-2002-bleu}, METEOR \cite{lavie-agarwal-2007-meteor}, and ROUGE-L \cite{lin-2004-rouge} as well as embedding based metrics: SkipThought \cite{NIPS2015_5950}, embedding average \cite{forgues2014bootstrapping}, vector extrema and Greedy Matching \cite{rus-lintean-2012-comparison}.
For reading ease, we reports metrics scaled between 0 and 100 rather than 0 and 1. 

\subsubsection{Automatic Diversity Evaluation}
We compute the type/token ratio on tokenized text, using the same spaCy\footnote{\url{https://spacy.io/}} tokenization used in the quality evaluation scripts.\footnote{\url{https://github.com/Maluuba/nlg-eval/tree/846166566bf0fdccbaa9e5b41da97147470b525b}}
\clearpage
\section{Extended Automatic Results}
\label{app:results}
\autoref{tab:multiref_word} and 
\autoref{tab:multiref_embed} show the evaluation against the multiple references for the word based and embedding based metrics. 
\autoref{tab:singleref_word} and 
\autoref{tab:singleref_embed} show the evaluation against the original single reference  for the word based and embedding based metrics. 
\begin{table*}
\centering
\addtolength{\tabcolsep}{-3pt}
\begin{adjustbox}{max width=\linewidth}
\begin{tabular}{l|cccc|cccc|cc}
\toprule
& \multicolumn{4}{c|}{Average Max Sentence BLEU} & \multicolumn{4}{c|}{Corpus BLEU} & \multicolumn{2}{c}{}  \\ 
               &BLEU1 & BLEU2 & BLEU3 & BLEU4 &BLEU1 & BLEU2 & BLEU3 & BLEU4& METEOR & ROUGE \\
               \midrule
baseline       & 27.9 & 14.3 & 9.8  & 7.3 & 48.3 & 25.1 & 15.3 & 10.0 & 12.8 & 34.0 \\
SMRT           & 29.2 & 16.4 & 11.6 & 8.9 & 49.9 & 28.1 & 18.1 & 12.4 & 13.8 & 36.1 \\\midrule
baseline + MMI & 27.8 & 13.8 & \phantom{1}9.3  & 7.0 & 48.2 & 24.3 & 14.6 & \phantom{1}9.5  & 12.7 & 33.5 \\
SMRT + MMI     & 29.2 & 16.2 & 11.5 & 8.7 & 50.1 & 27.9 & 17.9 & 12.2 & 13.7 & 35.8 \\\midrule
PT + baseline  & 29.5 & 15.9 & 11.0 & 8.3 & 49.9 & 27.1 & 16.9 & 11.3 & 13.6 & 35.8 \\
PT + SMRT      & 29.7 & 16.6 & 11.8 & 9.0 & 50.7 & 28.4 & 18.1 & 12.3 & 13.9 & 36.6\\
\bottomrule
\end{tabular}
    \end{adjustbox}

\caption{Word-overlap based metrics on multiple references.}
\label{tab:multiref_word}
\end{table*}

\begin{table*}[]
\addtolength{\tabcolsep}{-2pt}
\centering
\begin{tabular}{l|ccc|c}
\toprule
 & \multicolumn{3}{c|}{Cosine Similarity} & \\ 
               & SkipThought & Embed. Avg. &  VectorExtrema & GreedyMatching \\
               \midrule
baseline       & 71.7 & 90.6 & 62.2 & 76.9 \\
SMRT           & 73.6 & 90.5 & 63.4 & 77.7 \\\midrule
baseline + MMI & 71.6 & 90.7 & 62.3 & 76.7 \\
SMRT + MMI     & 73.5 & 90.5 & 63.3 & 77.6 \\\midrule
PT + baseline  & 72.5 & 90.9 & 63.2 & 77.5 \\
PT + SMRT      & 73.8 & 90.5 & 63.5 & 77.6\\
\bottomrule
\end{tabular}
\caption{Embedding based metrics on multiple references}
\label{tab:multiref_embed}
\end{table*}

\begin{table*}[]
\centering
\addtolength{\tabcolsep}{-3pt}
\begin{adjustbox}{max width=\linewidth}
\begin{tabular}{l|cccc|cccc|cc}
\toprule
& \multicolumn{4}{c|}{Average Max Sentence BLEU} & \multicolumn{4}{c|}{Corpus BLEU} & \multicolumn{2}{c}{}  \\ 
               &BLEU1 & BLEU2 & BLEU3 & BLEU4 &BLEU1 & BLEU2 & BLEU3 & BLEU4& METEOR & ROUGE \\
               \midrule
baseline       & 15.8 & \phantom{1}7.5  & 5.4 & 4.2 & 14.0 & 6.6 & 4.1 & 2.8 & 6.9 & 20.9 \\
SMRT           & 18.0 & 10.0 & 7.4 & 5.8 & 15.1 & 8.2 & 5.5 & 3.9 & 8.1 & 24.0 \\\midrule
baseline + MMI & 15.4 & \phantom{1}7.0  & 5.0 & 3.9 & 13.7 & 6.2 & 3.8 & 2.6 & 6.6 & 20.1 \\
SMRT + MMI     & 17.9 & \phantom{1}9.7  & 7.2 & 5.7 & 15.1 & 8.0 & 5.3 & 3.8 & 7.9 & 23.5 \\\midrule
PT + baseline  & 16.4 & \phantom{1}8.0  & 5.7 & 4.5 & 14.6 & 7.0 & 4.4 & 3.0 & 7.1 & 21.7 \\
PT + SMRT      & 17.9 & \phantom{1}9.8  & 7.3 & 5.7 & 15.2 & 8.1 & 5.3 & 3.8 & 7.9 & 23.7\\
\bottomrule
\end{tabular}
    \end{adjustbox}

\caption{Word-overlap based metrics on the single reference test set}
\label{tab:singleref_word}
\end{table*}

\begin{table*}[]
\addtolength{\tabcolsep}{-2pt}
\centering
\begin{tabular}{l|ccc|c}
\toprule
 & \multicolumn{3}{c|}{Cosine Similarity} & \\ 
               & SkipThought & Embed. Avg. &  VectorExtrema & GreedyMatching \\
               \midrule
baseline       & 64.8 & 86.1 & 50.0 & 71.2 \\
SMRT           & 67.2 & 86.5 & 52.1 & 72.5 \\\midrule
baseline + MMI & 64.6 & 86.0 & 49.9 & 70.8 \\
SMRT + MMI     & 67.0 & 86.4 & 51.9 & 72.3 \\\midrule
PT + baseline  & 65.4 & 86.4 & 50.6 & 71.5 \\
PT + SMRT      & 67.1 & 86.5 & 52.0 & 72.3\\
\bottomrule
\end{tabular}
\caption{Embedding based metrics on the single reference test set}
\label{tab:singleref_embed}
\end{table*}
\clearpage

\section{Examples}
\label{app:ex}
\subsection{Paraphrase Examples}
\autoref{tab:para_example_house} and \autoref{tab:para_example_video} 
each show a training pair and 20 independent random paraphrases of the response. 
Sampling is limited to the top 100 tokens per time-step. During training a new sample is taken in each of the 100 epochs. While there are a few small errors, overall the paraphrases are of high quality and remain valid responses. Since sampling is redone each epoch, an error would only be seen once in training. 

\subsection{Dialog Examples}
Tables \ref{tab:dialog_example_kebab} through \ref{tab:dialog_example_spanish} show example evaluation prompts and example outputs. 

\begin{table*}
\centering
\addtolength{\tabcolsep}{-3pt}
\begin{tabular}{ll}
\toprule
prompt:&It's a wonderful Spanish style.
  \\
response:&Oh, I love the roof tiles on Spanish style houses. \\
  \midrule
paraphrases:&Ahmed says he loves the cladding of Spanish-style houses.\\
&Oh, I like the roof tiles on houses in Spanish style.\\
&Oh, I love the roof tiles of Spanish style houses.\\
&Oh, I love the roof tiles on Spanish-style houses.\\
&Oh, I love roof tiles on Spanish-architecture homes.\\
&Oh, I enjoy roof tiles, in Spanish-style homes.\\
&Oh, I love the roof tiles on Spanish-American style houses.\\
&Oh, I love the roof tills on Spanish style houses.\\
&Oh, I love the roof tiles in Spanish-style house.\\
&Oh, I love the roof tiles to house in Spanish style\\
&OH, I love the tiled roofs that are found in Spanish style homes.\\
&Oh, I love the roof tiles on Spanish-style houses.\\
&Oh, I like roof tiles on Spanish-style houses.\\
&Oh, I love the roof tiles on Spanish-style houses.\\
&Oh I'm very fond of roof tiles in Spanish-style homes.\\
&Oh, I adore the roof tiles on Spanish-style homes.\\
&Oh, I love the roof tiles on Spanish style houses.\\
&Oh, I love roof tiles in Spanish-style houses.\\
&Oh, I love roof tilting on Spanish-style houses.\\
&Oh, I love the roof tiles on Spanish style houses.\\
\bottomrule
\end{tabular}
\caption{An example training pair with 20 independent random paraphrase samples of the response. Sampling is limited to the top 100 tokens per time-step. During training a new sample is taken each epoch.}
\label{tab:para_example_house}
\end{table*}

\begin{table*}
\centering
\addtolength{\tabcolsep}{-3pt}
\begin{tabular}{ll}
 \toprule
prompt:&Oh, man. I hope you get to feeling better.\\
response:&I would feel better if you came over and took care of me.\\\midrule
paraphrases:&I'd be well if you came up and took care of me.\\
&I would feel better if you came down and took care of me.\\
&I would be better off if you came up and took care of me.\\
&And I wish you could make care of me and find me.\\
&I would do better if you came and took care of me.\\
&I'd feel better if You came up and took care of me.\\
&I would feel better if you came up and took care of me.\\
&I would feel better if you come over and take care of me.\\
&I will feel best if you come over and take care of me.\\
&I would feel better if you came and took care of me.\\
&This is something I would feel better if you came up and took care of me.\\
&I would feel better if you came and cared for me.\\
&I would feel better if you came up and took care of me.\\
&I would feel better if you came up and took care of me.\\
&I would feel better if you came up and took care of me.\\
&I would feel better if you came out and took care of myself.\\
&I'd be better off if you came over and took care of me.\\
&I hope you'd grow up and take care of me.\\
&I would feel better if you got over and looked after me.\\
&I would feel better if you came down and picked me up.\\\bottomrule
\end{tabular}
\caption{An example training pair with 20 independent random paraphrase samples of the response. Sampling is limited to the top 100 tokens per time-step. During training a new sample is taken each epoch.}
\label{tab:para_example_video}
\end{table*}

\begin{table*}
 \centering
\addtolength{\tabcolsep}{-3pt}
\begin{tabular}{ll}
\toprule
prompt:&that sounds nice. Do you want\\& to share some kebabs as well?\\
\midrule
baseline:&sure. How about some cookies?\\
SMRT:&sure, what about some lambs?\\
PT + baseline:&yes, I'd like to share some kebabs.\\
PT + SMRT:&yes, I would like to share some kebabs.\\
\bottomrule
\end{tabular}
\caption{An evaluation prompt and example output. The baseline mentions the wrong kind of food. The remaining systems give good responses, though perhaps SMRT should be suggesting `lamb' (kebab) rather than `lambs'.}
\label{tab:dialog_example_kebab}
\end{table*}

\begin{table*}
 \centering
\addtolength{\tabcolsep}{-3pt}
\begin{tabular}{ll}
\toprule
prompt:&I'll use ginger, garlic scallion, hot pepper and vinegar.\\
\midrule
baseline:&I'm afraid I'm not sure.\\
SMRT:&I'm sorry, I'm going to buy it!\\
PT + baseline:& Anything else?\\
PT + SMRT:&That sounds like a good idea.\\
\bottomrule
\end{tabular}
\caption{An evaluation prompt and example output. The baseline is off-topic, but SMRT is better here. Baseline and SMRT pretraining each give good responses.}
\label{tab:dialog_example_ginger}
\end{table*}

\begin{table*}
 \centering
\addtolength{\tabcolsep}{-3pt}
\begin{tabular}{ll}
\toprule
prompt:&Believe it or not, tea is the most popular beverage in the world after water.\\
\midrule
baseline:&Really? What's the nature of this city?\\
SMRT:&Really? I am so surprised.\\
PT + baseline:& That's true. Tea is the most popular in the world.\\
PT + SMRT:& Yes, I agree with you, but I don't know what to do.\\
\bottomrule
\end{tabular}
\caption{An evaluation prompt and example output. The baseline is off-topic, but SMRT and baseline pretraining give good (but very different) responses. Pretrained SMRT is less ideal, though could be a fine response in some contexts. }
\label{tab:dialog_example_tea}
\end{table*}

\begin{table*}
 \centering
\addtolength{\tabcolsep}{-3pt}
\begin{tabular}{ll}
\toprule
prompt:&Oh, good. How old is your niece?\\
\midrule
baseline:&I'm afraid I'm not sure.\\
SMRT:&Eighteen.\\
PT + baseline:&She's 42.\\
PT + SMRT:&She is ten years old.\\
\bottomrule
\end{tabular}

\caption{An evaluation prompt and example output. The baseline gives an `I don't know' variant, but the remaining systems have good output.}
\label{tab:dialog_example_neice}
\end{table*}

\begin{table*}
 \centering
\addtolength{\tabcolsep}{-3pt}
\begin{tabular}{ll}
\toprule
prompt:&Can you tell me how to get to the Jade Buddha Temple?
\\
\midrule
baseline:&Sure. You can buy yourself a ID card.\\
SMRT:&You can buy a ticket from the buyer's cheque.\\
PT + baseline:&Sure. Go straight ahead and turn left.\\
PT + SMRT:&Go straight ahead and turn left, turn left, turn left and turn left.\\
\bottomrule
\end{tabular}
\caption{An evaluation prompt and example output. The baseline does not answer the question, but the remaining systems have good output.}
\label{tab:dialog_example_temple}
\end{table*}

\begin{table*}
 \centering
\addtolength{\tabcolsep}{-3pt}
\begin{tabular}{ll}
\toprule
prompt:&I know what you mean. I ought to wear right clothes at the right time.
\\
\midrule
baseline:&I don't know. I'm not sure what to do.\\
SMRT:&I guess you are right.\\
PT + baseline:&I'm sorry to hear that. I don't know what to wear.\\
PT + SMRT:&I don't know what to wear at the right time.
\\
\bottomrule
\end{tabular}
\caption{An evaluation prompt and example output. The baseline gives an `I don't know' variant. While the pretrained baseline has some uncertainty, it is at least unsure about clothing. The SMRT systems have good output.}
\label{tab:dialog_example_clothes}
\end{table*}

\begin{table*}
 \centering
\addtolength{\tabcolsep}{-3pt}
\begin{tabular}{ll}
\toprule
prompt:&Do you speak only English in the class, \\
&or does your teacher explain everything to you in Spanish?
\\
\midrule
baseline:&No, I don't know how to speak Spanish.\\
SMRT:&Oh, we never speak Spanish.\\
PT + baseline:&I don't know.\\
PT + SMRT:&No, I don't speak English in the class.
\\
\bottomrule
\end{tabular}
\caption{An evaluation prompt and example output. The baseline pretraining gives an `I don't know' variant. The baseline is okay, the SMRT systems have better output.}
\label{tab:dialog_example_spanish}
\end{table*}

\end{document}